# Fine Tuning Large Language Models for Medicine:
# The Role and Importance of Direct Preference Optimization

Thomas Savage MD[1,2], Stephen Ma MD PhD[1,2], Abdessalem Boukil BS[3], Vishwesh Patel MBBS[4], Ekanath Rangan MD[1], Ivan Lopez BS[6], Jonathan H Chen MD PhD[1,2,5,6]

[1] Department of Medicine, Stanford University, Stanford, CA
[2] Division of Hospital Medicine, Stanford University, Stanford, CA
[3] Linguamind AI
[4] M.P. Shah Government Medical College, Gujarat, India
[5] Stanford Center for Biomedical Informatics Research, Stanford University, Stanford, CA
[6] Clinical Excellence Research Center, Stanford University, Stanford, CA

Corresponding Author: Thomas Savage

## Abstract

Large Language Model (LLM) fine tuning is underutilized in the field of medicine.  Two of the most common methods of fine tuning are Supervised Fine Tuning (SFT) and Direct Preference Optimization (DPO), but there is little guidance informing users when to use either technique.  In this investigation, we compare the performance of SFT and DPO for five common natural language tasks in medicine: Classification with text data, Classification with numeric data, Clinical Reasoning, Summarization, and Clinical Triage.  We find that SFT alone is sufficient for Classification with text data, whereas DPO improves performance for the more complex tasks of Clinical Reasoning, Summarization and Clinical Triage.  Our results establish the role and importance of DPO fine tuning within medicine, and consequently call attention to current software gaps that prevent widespread deployment of this technique.

## Introduction

The initial lofty expectations of large language models (LLMs) in medicine have entered a lull, where the field is beginning to realize that deploying out-of-the-box generalist models may not achieve sufficient accuracy to be immediately impactful.[1,2]  Researchers have begun to experiment with different strategies such as prompt engineering and Retrieval Augmented Generation to improve model performance[3–5], but one performance strategy that has not received sufficient attention by the field of medicine is fine tuning.

Medicine has been slow to adopt fine tuning due to multiple contributing factors.  First, there is an over-reliance on closed source models that have limited fine tuning functionality.[6–8]  Second there is a widely held misconception that fine tuning requires prohibitively large amounts of example data that is impractical for a single hospital or clinic to collect within the data siloes of medicine.[9]  And finally there is a scarcity of literature informing users what Natural Language Processing (NLP) tasks benefit from LLM fine tuning, and if so, which specific fine tuning methods should be deployed.   Therefore, in this study we aim to quantify the benefits of two fine tuning techniques: Supervised Fine Tuning (SFT) and Direct Preference Optimization (DPO), across key elementary tasks in clinical NLP with small datasets (less than 6,000 examples).

Fine tuning is the process of adjusting the coefficient weights of a language model after pre-training, adapting the model with a subject-specific dataset of interest to the user.[9–12] Two of the most common methods for fine-tuning language models are Supervised Fine Tuning (SFT) and Direct Preference Optimization (DPO).

Supervised Fine Tuning (SFT) is the traditional method of fine tuning a language model. SFT requires the user to provide example prompts and gold standard responses.  SFT uses a classic loss function to adjust model weights and maximize the probability the model will re-produce similar gold standard responses.[13]  In many ways SFT is simply training the model to mimic the gold standard responses.

Direct Preference Optimization (DPO) is a variation of Reinforcement Learning (RL) that has become a popular fine tuning technique because of its stability when training with smaller datasets.[14]  In comparison to SFT, DPO requires the user to provide not only prompts and gold standard responses, but also "rejected" answers that the user finds undesirable.  The use of rejected answers for fine tuning is the key difference between SFT and DPO, because DPO adjusts model weights to both maximize the likelihood of gold standard responses, while also minimizing the likelihood of rejected responses.  This conceptual difference is reflected in the DPO loss function (Supplemental Information I).[9,13,14]

When to use DPO over SFT is an area of active investigation.   DPO is described as providing better alignment with human preferences, but recent publications have highlighted the ambiguity of this description.[13]  It is unknown whether better alignment translates to better reasoning, summarization, information retrieval, or other tasks of importance to clinicians.  Overall few studies have compared SFT against DPO for individual NLP tasks[9], and fewer have evaluated fine tuning for NLP tasks specific to medicine.

In this study we aimed to establish what clinical NLP tasks benefit from SFT and DPO.  We selected five elemental clinical NLP tasks to evaluate from the recent systematic review by Bedi et al.[15]  We investigated Classification with text data, Classification with numeric data, Clinical Triage, Clinical Reasoning (diagnosis and treatment selection), and Clinical Text Summarization.  Table 1 details the dataset used to evaluate each clinical NLP task. We performed our experiment on two popular open source LLMs: Llama3-8B-Instruct[16] and Mistral-Instruct-v2[17], using datasets of less than 6,000 training examples.

| Task | Description | Clinical Scenario Tested | Dataset | Gold Standard Answer | Rejected Answer |
|---|---|---|---|---|---|
| Classification (Text Data) | Recognize a strict text-based criterion, classifying a passage into one of multiple groups. | Identify passages describing patients with a Urinary Tract Infection (pyuria with lower urinary tract symptoms) versus only pyuria. | Total Dataset Size: 700<br><br>Patient scenarios were generated by GPT-4[18] and then edited by three board certified physicians for accuracy and sufficient variability. | Diagnosis by board certified physician. | Incorrect diagnosis not selected by grading physician. |
| Classification (Numeric Data) | Recognize a numeric-based criterion, classifying a passage into one of multiple groups. | Interpret urine chemistries for a patient with hyponatremia. Diagnose either Hypovolemic Hyponatremia, Primary Polydipsia or Syndrome of Inappropriate Anti-Diuretic Hormone Secretion (SIADH). | Total Dataset Size: 1,200<br><br>Patient scenarios were syntactically identical except for randomly generated lab values for urine sodium, urine osmolality and serum sodium. | Diagnosis by board certified physician. | Incorrect diagnoses not selected by grading physician. |
| Clinical Triage | Recognize abstract criterion to classify a passage into one of multiple groups. | Triage patient messages for both the appropriate urgency of response (urgent or non-urgent) and appropriate responding provider (physician or medical assistant). | Total Dataset Size: 1,800<br><br>Outpatient clinic patient messages from Stanford Healthcare triaged by physician author TRS according to criteria listed in Supplemental Information VII | Appropriate triage as determined by the grading physician. | Incorrect triage not selected by the grading physician. |
| Clinical Reasoning | Interpret patient information to identify diagnoses and select treatments. | Medical board exam questions evaluating the skills of clinical diagnosis and treatment selection. | Total Dataset Size: 5,161<br><br>MedQA dataset, modified to questions evaluating clinical diagnosis and treatment selection at the STEP 2 and 3 level. | Correct answer provided by the MedQA dataset. | Randomly selected incorrect multiple-choice option provided by the MedQA dataset. |
| Summarization | Identify key information in a passage for a target audience. | Summarize a discharge summary note into 2-3 sentences for an internal medicine physician. | Total Dataset Size: 5,250<br><br>Synthetic discharge notes from the AISC Augmented Clinical Notes dataset.[19] | GPT-4[18] generated summaries. | Llama2[20] Generated Summaries. |

Table 1. Description of the NLP tasks evaluated and corresponding dataset, gold standard answer, and rejected answer. All datasets (except for patient message triage) are provided in Supplemental Information II-VI.

# Results

### Classification using Text Data
The first elementary task studied, Classification with text data, found base Llama3 and Mistral2 to achieved F1 scores of 0.63 and 0.73 respectively when identifying passages describing patients with a Urinary Tract Infection (UTI). When trained with SFT, Llama3's F1 score increased to 0.98 while Mistral2 increased to 0.97. DPO fine tuning decreased Llama3's F1 score to 0.95 (p value 0.55 when compared to SFT) and did not change Mistral2, holding at 0.97 (p value 1.0 when compared to SFT). Results are provided in Figure 1a.

### Classification using Numeric Data
The second elementary task, Classification with numeric data, found both base Llama3 and Mistral2 with an F1 score of 0.18 when interpreting urine electrolyte laboratories for patients with hyponatremia. When fine-tuned with SFT, Llama3 F1 score remained at 0.18 and Mistral2's F1 score decreased to 0.16. When fine tuned with DPO, Llama3 F1 score increased to 0.27 (p value 0.79) and Mistral2 remained at 0.16 (p value 1.0). Results are provided in figure 1b.

### Clinical Reasoning
The third elementary task, Clinical Reasoning, found base Llama3 and Mistral achieved accuracies of 7% and 22% respectfully on a modified MedQA dataset. SFT modeling increased accuracy to 28% and 33% respectively. DPO increased accuracy even further to 36% (p value 0.003) for Llama3 and 40% (p value 0.004 ) for Mistral2. Results are illustrated in Figure 1c.

### Clinical Summarization
The fourth elementary task evaluated was Summarization, evaluating whether each model could effectively condense clinical notes into a 2-3 sentence summary. Summaries produced by base Llama3 achieved an average Likert rating of 4.11 and base Mistral summaries achieved a rating of 3.93 (a rating of five was the highest score and one was the lowest). SFT fine tuning improved ratings to 4.21 for Llama3 and 3.98 for Mistral2. DPO even further improved ratings to 4.34 (p value <0.001) for Llama3 and 4.08 (p value <0.001) for Mistral2. Results are shown in Figure 1d.

### Clinical Triage
The final elementary task evaluated was Clinical Triage, where models were fine tuned to triage patient messages from Stanford Healthcare Clinics for both the appropriate responding provider (medical assistant or physician) and urgency (urgent vs non-urgent). Base Llama3 achieved F1 scores of 0.55 and 0.81 for personnel and urgency triage respectively, while base Mistral2 achieved F1 scores of 0.49 and 0.88. SFT fine tuning increased Llama3's F1 score to 0.58 for personnel triage but decreased its F1 score for urgency triage to 0.79. Mistral2 with SFT increased its personnel triage F1 score to 0.58 and decreased its urgency triage F1 score to 0.87. DPO increased Llama3 performance for both personnel to 0.74 (p value <0.001) and urgency triage: 0.91 (p value <0.001). Mistral2 experienced a similar improvement with DPO for personnel triage to 0.66 (p value <0.001) but did not experience benefit for urgency triage, with an F1 score of 0.85 (p value 1.0). Figures 1e and 1f show F1 score results, while sensitivity and specificity data are provided in Supplemental Tables 2 and 3.

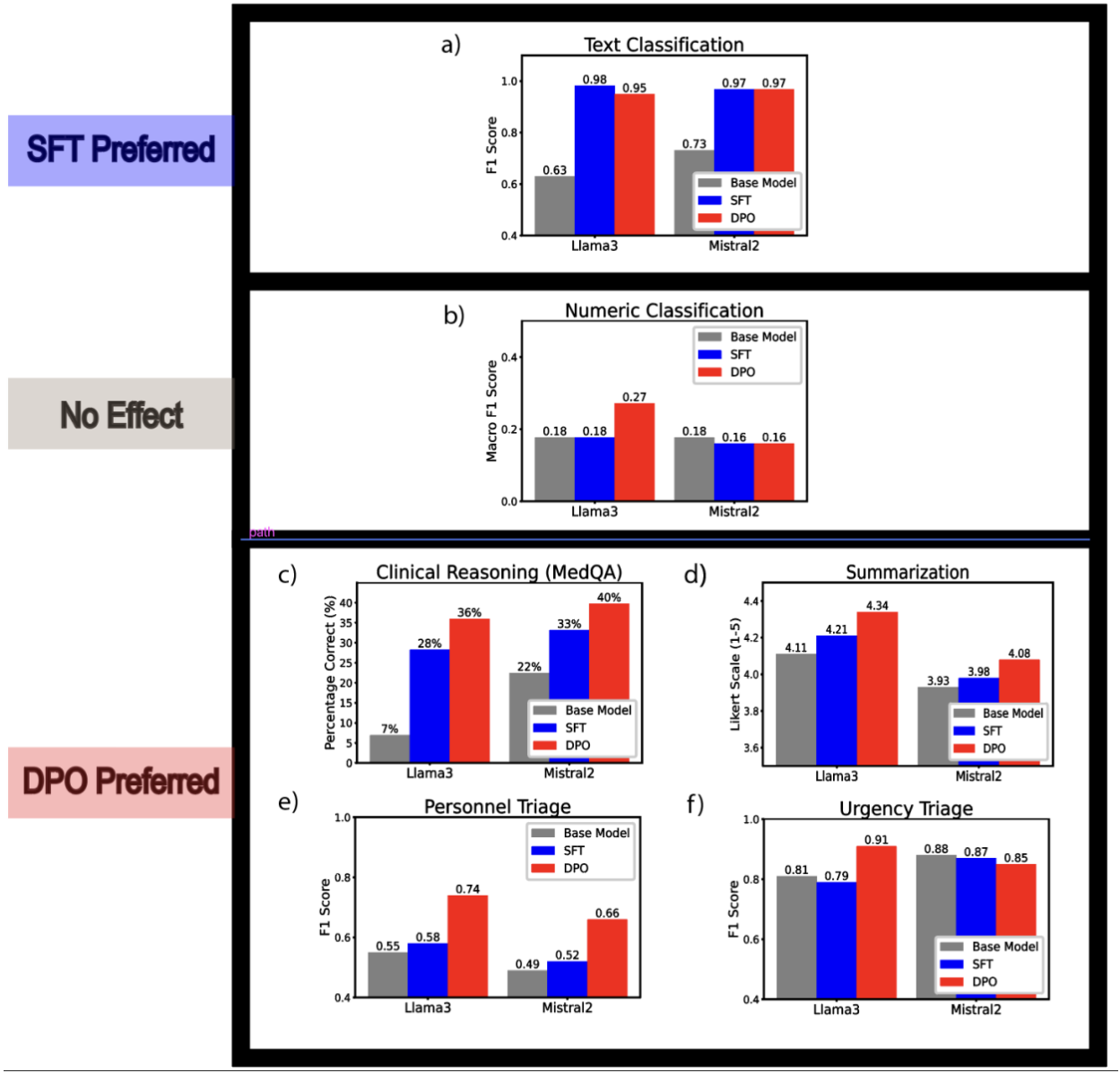

*Figure 1. Comparison of base Llama3 and Mistral2 (gray) against SFT (blue) and DPO (red) fine-tuned variants for the tasks of Text Classification, Numeric Classification, Clinical Reasoning, Triage and Summarization.*

## Discussion

The results of our investigation identify when fine tuning with Supervised Fine Tuning and Direct Preference Optimization benefit common clinical natural language tasks. We found SFT alone was sufficient for text-based Classification (Figure 1a), while the more complex tasks of Triage, Clinical reasoning, and Summarization showed statistically significant improvement with DPO (Figures 1c, 1d, 1e, and 1f). In contrast, neither fine tuning method was beneficial in performing numeric-based Classification (Figure 1b).

We hypothesize that SFT alone is sufficient for text-based Classification but not for Triage, Clinical Reasoning, or Summarization because SFT strengthens simple "word-association" reasoning, while DPO enables deeper interpretation. Because SFT is trained on only gold-standard responses, the model is conditioned to recognize high yield words or basic concepts, but not deeper comprehension. By comparison, DPO is trained with both positive and negative examples, and this contrast enables the model to recognize more complex patterns (mimicking better understanding). As a result we observe SFT alone is sufficient for Classification tasks with clearly defined criteria, such as diagnosing a UTI, whereas DPO fine-tuning is better for Classification tasks that have abstract criteria such as patient message triage or summarization. Therefore we conclude that SFT alone is sufficient for simple tasks that can be achieved by word or entity association, whereas DPO is superior for tasks that require recognition of more advanced patterns.

Our findings demonstrate the importance of fine tuning with DPO, and we believe it (along with its variants[9]) will play a large role in the future of clinical natural language processing. Because the medical community is functionally partitioned, with unique practice and documentation styles across health systems, the field will demand a high degree of personalization from language models. SFT and DPO will be required to learn the specialized knowledge and physician preferences that are unique to each health system.

Nevertheless, before DPO can be more widely applied, the informatics community must address software gaps that block its widespread deployment. First, most closed source language model vendors, such as OpenAI and Anthropic, do not seem to offer DPO fine tuning.[21,22] Their exact methods are not disclosed, but neither fine-tuning API accepts "rejected" samples, suggesting against a traditional DPO technique. This prevents many leading models such as GPT-4 or Claude-3 from being sufficiently optimized for clinical applications. Second, current open-source python DPO libraries lack the intrinsic ability to parallelize between multiple GPUs[23], preventing full precision fine tuning of LLMs larger than ten billion parameters. This discourages utilization of DPO amongst informaticists unfamiliar with distributed system training, requiring the user to write custom code. Ultimately, to facilitate widespread adoption of DPO with our strongest models, the Generative AI community must work together to lift these obstacles.

A limitation of our investigation is we did not evaluate language models larger than ten billion parameters. We expect our findings will generalize to larger models but were unable to complete such an experiment due to cost. The cost of fine tuning one model of size 7B on a single A100 GPU is approximately 12 GPU hours, whereas a 70B model would require at least 16 GPUs and significantly higher costs to perform full precision fine tuning. In turn, we believe our robust exploration of smaller models provides valuable insight to guide investment in fine tuning larger models that will be used in clinical operations or care.

A notable strength of our investigation is the use of small datasets smaller than 6,000 examples to reflect the data limitations of clinical medicine. Many existing publications on fine tuning deploy training sets greater than 30,000 examples[9,24–26], sizes that are unrealistic for a single hospital system or clinic to achieve. Therefore, our findings prove the feasibility of fine-tuning language models within the realistic data constraints of medicine.

## Conclusions

Fine tuning with SFT alone is sufficient for text-based Classification with well-defined criteria. In contrast, fine tuning with DPO optimizes performance for more complex tasks with abstract criteria such as Triage, Clinical Reasoning and Summarization.

Before DPO can be widely deployed in medicine, software gaps must be addressed. Closed source models must offer DPO functionality and open source python DPO libraries must facilitate parallelization between GPUs.

## Methods

### Overview

SFT and DPO were compared over five datasets, each evaluating one of five elementary NLP tasks important to medicine. Each dataset consisted of a train, evaluation, development, and test set. The base LLM model was first fine-tuned via SFT using the train and evaluation datasets, and then the development dataset was used to select the top performing SFT model. The top performing SFT model was then used as the base model for DPO fine-tuning. DPO was then performed using the train and evaluation datasets, and the top performing DPO model was selected using the development set. Finally the base LLM, top performing SFT model, and top performing DPO model were compared using the test set. This evaluation process is illustrated in Figure 2.

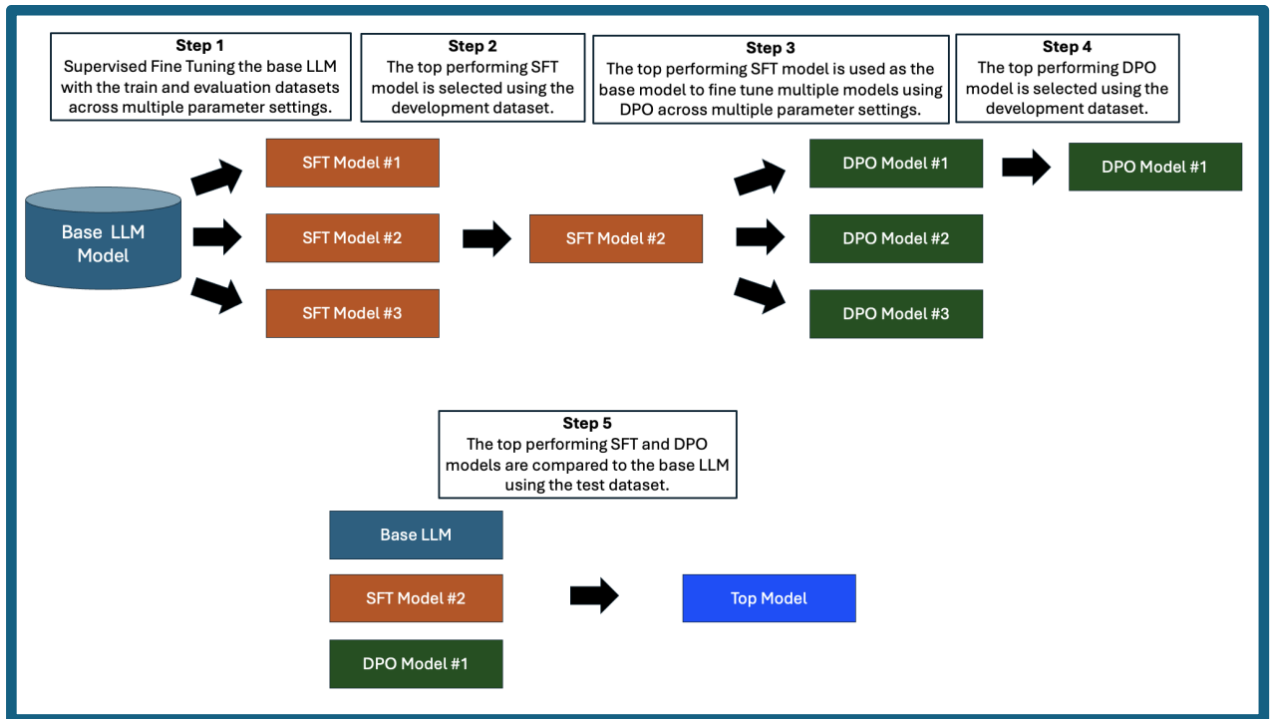

*Figure 2. Overview of study methods fine tuning SFT and DPO models, and comparing those models to the base LLM.*

## Elementary Tasks Evaluated

### Classification with Text Data
The first elementary task evaluated was Classification with text data, where models were asked to identify passages describing patients with a possible urinary tract infection (UTI). To be classified as a UTI, the passage needed to describe both pyuria and lower urinary tract symptoms.

The dataset was generated by GPT-4, which was prompted to generate 400 cases describing pyuria with no symptoms and 400 cases describing pyuria with urinary symptoms (positive for UTI). Three physician annotators then reviewed the generated cases to ensure correctness and introduce sufficient variability between examples. The 800 examples were then split into a training set (300 examples), evaluation set (200 examples), dev set (100 examples) and test set (200 examples). Prompts, patient descriptions, and model responses with grades are provided in Supplemental Information II.

### Classification with Numeric Data
The second elementary task evaluated was Classification with numeric data, where the models were asked to interpret urine electrolyte studies in patients with hyponatremia. The model was asked to recognize hypovolemic hyponatremia versus hyponatremia secondary to primary polydipsia versus hyponatremia secondary to Syndrome of Inappropriate Antidiuretic Hormone Section (SIADH).

The dataset comprised of 1,200 example cases that contained the exact same text, differing only by their values of Urine Sodium, Urine Osmolality and Serum Sodium (Supplemental Information III). Numeric laboratory values were randomly generated to be consistent with either hypovolemia, primary polydipsia, or SIADH, with equally distributed prevalence. The dataset consisted of 700 training examples, 200 evaluation examples, 100 development examples, and 200 test examples. Model responses and grades can be found in Supplemental Information III.

### **Clinical Reasoning**
The third elementary task evaluated was Clinical Reasoning. Clinical reasoning was evaluated using a modified MedQA dataset, where the original MedQA dataset was adapted to have open ended questions and only include STEP 2 and 3 level board exam questions (assessments that focus on higher levels of clinical reasoning).

The modified MedQA dataset consisted of 4095 training examples, 456 evaluation examples, 200 development examples, and 410 test questions. Gold-standard answers were identified as the original MedQA answer, and rejected answers (used for DPO fine tuning) were randomly selected from the list of incorrect multiple-choice options from the original dataset.

Each open-ended question was graded by at least two physician annotators (authors ER, VP, and TS). If there was disagreement over the grade by the first two physician annotators, the third annotator decided the final grade. Full data along with model responses graded by can be found in Supplemental Information IV.

### **Summarization**
The fourth elementary task evaluated was Summarization, where the models were asked to summarize discharge summaries into 2-3 sentences. Synthetic discharge summary notes were taken from the AISC Augmented Clinical Notes dataset.[19] Gold standard summaries were generated by GPT-4 (gpt-4-0613)[18] and rejected examples for DPO training were generated by the Llama2-chat-7B model.[30] Code used to generate training summaries can be found in Supplemental Information V.

The AISC dataset consisted of 4,500 training examples, 300 evaluation examples, 150 development examples, and 300 test examples. LLM summaries were judged by GPT-4 on a Likert scale of one to five, with five being the best possible score. Full data along with model grades can be found in Supplemental Information VI.

### **Triage**
The fifth elementary task evaluated was Triage, where the model was asked to triage patient messages for appropriate urgency (urgent vs non-urgent) and the appropriate responding provider (medical assistant versus physician). Patient messages were sourced from Stanford Clinics and graded by author TRS. Criteria for grading is provided in Supplemental Information VII.

2,400 total messages were graded. Messages that were ambiguous or did not require a response were not included in our experiment. The final dataset consisted of 1,300 training examples, 200 evaluation examples, 100 development examples, and 200 test examples.

**Fine Tuning Hyperparameters**
Optimal hyperparameters were tested for with a sweep, and the optimal settings determined by testing on the dev set. Learning rates tested were $10^{-5}$, $10^{-6}$, $10^{-7}$, and $10^{-8}$. Beta values tested were 0.1, 0.3 and 0.5.

Each model-hyperparameter configuration was initially tested with 1,000 steps. The validation error plot was then analyzed to identify where the model plateaued, and then the model was trained a second time with that step count.

All models produced by this experiment are available at the huggingface account *tsavage68*.

**Statistical Evaluation**
A McNemar's test was used for statistical evaluation of tasks with binary outcomes (Classification with text data, Classification with numeric data, Clinical Reasoning, and Triage). A paired T test was used for statistical evaluation of tasks with ordinal outcomes (Summarization). An alpha of 0.01 was used as our statistical significance threshold, accounting for Bonferroni correction.

**Supplemental Information I**
Direct Preference Optimization loss function.

$$\nabla_\theta L_{DPO}(\pi_\theta; \pi_{ref}) = -\beta \mathbb{E}_{(x,y_w,y_l) \sim D} \left[ \sigma(\hat{r}_\theta(x, y_l) - \hat{r}_\theta(x, y_w)) \ [\nabla_\theta \log \pi(y_w \mid x) - \nabla_\theta \log \pi(y_l \mid x)] \right]$$

**Supplemental Information II**
UTI Example Files.

**Supplemental Information III**
Hyponatremia Example Files.

**Supplemental Information IV**
Clinical Reasoning Example Files.

**Supplemental Information V**
Python code used to generate Information Prioritization examples with GPT-4 and Llama2.

**Supplemental Information VI**
Information Prioritization Example Files.

**Supplemental Information VII**
Triage criteria and prompt.

**Supplemental Information VIII**
SFT and DPO code as well as code for DPO with GPU Parallelization.

|  |  | Sensitivity | Specificity |
|---|---|---|---|
| Llama3 | Base | 30% | 99% |
|  | SFT | 34% | 99% |
|  | DPO | 80% | 84% |
|  |  |  |  |
| Mistral2 | Base | 20% | 94% |
|  | SFT | 25% | 95% |
|  | DPO | 64% | 89% |

Supplemental Table 2. Sensitivity and specificity for each base model with and without fine tuning for triaging patient messages appropriate for a physician (versus medical assistant).

|  |  | Sensitivity | Specificity |
|---|---|---|---|
| Llama3 | Base | 89% | 70% |
|  | SFT | 76% | 69% |
|  | DPO | 63% | 90% |
| Mistral2 | Base | 63% | 85% |
| | SFT | 64% | 83% |
| | DPO | 76% | 78% |

Supplemental Table 3. Sensitivity and specificity for each base model with and without fine tuning for triaging patient messages to be marked urgent (versus non-urgent).